\documentclass[conference]{IEEEtran}
\IEEEoverridecommandlockouts

\usepackage{cite}
\usepackage{amsmath,amssymb,amsfonts}
\usepackage{algorithmic}
\usepackage{graphicx}
\usepackage{textcomp}
\usepackage{xcolor}
\usepackage{bm}
\usepackage{mathtools, cuted}
\def\BibTeX{{\rm B\kern-.05em{\sc i\kern-.025em b}\kern-.08em
    T\kern-.1667em\lower.7ex\hbox{E}\kern-.125emX}}

\usepackage{amsthm}

\newtheorem{Lemma}{Lemma}

\newtheorem{Remark}{Remark}

\newcommand{\R}{\mathbb{R}}

\newcommand{\Imax}{\mathcal{I}_{\max}}
\newcommand{\Lmax}{\mathcal{L}_{\max}}
\newcommand{\Smax}{\mathcal{S}_{\max}}
\newcommand{\Lset}{\mathcal{L}_T}
\newcommand{\Iset}{\mathcal{I}_T}

\allowdisplaybreaks[3]

\usepackage[dvipsnames]{xcolor}
\usepackage{hyperref}
\hypersetup{colorlinks,linkcolor={BrickRed},citecolor={blue},urlcolor={blue}}

\begin{document}

\title{Variable Splitting Binary Tree Models Based on Bayesian Context Tree Models \\for Time Series Segmentation
}

\makeatletter
\newcommand{\linebreakand}{%
  \end{@IEEEauthorhalign}
  \hfill\mbox{}\par
  \mbox{}\hfill\begin{@IEEEauthorhalign}
}
\makeatother

\author{
\IEEEauthorblockN{Yuta Nakahara}
\IEEEauthorblockA{\textit{Center for Data Science} \\
\textit{Waseda University}\\
Tokyo, Japan \\
y.nakahara@waseda.jp}
\and
\IEEEauthorblockN{Shota Saito}
\IEEEauthorblockA{\textit{Faculty of Informatics} \\
\textit{Gunma University}\\
Gunma, Japan \\
shota.s@gunma-u.ac.jp}
\and
\IEEEauthorblockN{Kohei Horinouchi}
\IEEEauthorblockA{\textit{Dept. of Applied Mathematics} \\
\textit{Waseda University}\\
Tokyo, Japan \\
horinochi\_18@toki.waseda.jp}
\and
\IEEEauthorblockN{Koshi Shimada}
\IEEEauthorblockA{\textit{Dept. of Applied Mathematics} \\
\textit{Waseda University}\\
Tokyo, Japan \\
i\_am\_koshi@suou.waseda.jp }
\linebreakand 
\IEEEauthorblockN{Naoki Ichijo}
\IEEEauthorblockA{\textit{Dept. of Applied Mathematics} \\
\textit{Waseda University}\\
Tokyo, Japan \\
1jonao@fuji.waseda.jp}
\and
\IEEEauthorblockN{Manabu Kobayashi}
\IEEEauthorblockA{\textit{Center for Data Science} \\
\textit{Waseda University}\\
Tokyo, Japan \\
mkoba@waseda.jp}
\and
\IEEEauthorblockN{Toshiyasu Matsushima}
\IEEEauthorblockA{\textit{Dept. of Applied Mathematics} \\
\textit{Waseda University}\\
Tokyo, Japan \\
toshimat@waseda.jp}
}

\maketitle

\begin{abstract}
We propose a variable splitting binary tree (VSBT) model based on Bayesian context tree (BCT) models for time series segmentation. Unlike previous applications of BCT models, the tree structure in our model represents interval partitioning on the time domain. Moreover, interval partitioning is represented by recursive logistic regression models. By adjusting logistic regression coefficients, our model can represent split positions at arbitrary locations within each interval. This enables more compact tree representations. For simultaneous estimation of both split positions and tree depth, we develop an effective inference algorithm that combines local variational approximation for logistic regression with the context tree weighting (CTW) algorithm. We present numerical examples on synthetic data demonstrating the effectiveness of our model and algorithm.
\end{abstract}


\section{Introduction}

In this study, we address the problem of time series segmentation, which is also called offline change point detection. It is the problem of detecting all the past change points of the probabilistic data generative models after observing all the data points. In this paper, the number of model changes is assumed to be unknown. Such problems are important and have been studied across a wide range of fields, e.g., financial analysis, bio-informatics, and climatology (see e.g., \cite{truong2020selective}).

Approaches to time series segmentation can be broadly classified into two categories. In the first approach, probabilistic models are assumed not only on the data generation processes but also on the change process of them. In many cases, such studies assume that changes occur independently at each time point with a certain probability, e.g., hidden Markov models \cite{HMM}, i.p.i.d. sources \cite{ipid}, and changing context tree models \cite{ChangingContextTree, kontoyiannis_cange_point}. One advantage of this approach is that the uncertainty of changes can be quantified by posterior probabilities.

In the second approach, probabilistic models are not assumed on the change process of the probabilistic data generative models. Change points are just detected using some algorithm. While this approach makes it more difficult to statistically evaluate the uncertainty of changes, it allows the use of more flexible algorithms. For example, algorithms that recursively repeat binary splits under some stopping rule have been studied (see e.g., \cite{truong2020selective}). The output of such algorithms is represented by a binary tree.

Traditionally, no probabilistic model representing the change process corresponding to such recursive binary splitting algorithms had been proposed. However, in recent years, a fixed splitting binary tree (FSBT) model \cite{Nakahara_ICSDS} where the tree itself is a random variable and represents the change process has been proposed by applying Bayesian Context Tree (BCT) models. The BCT model is a Bayesian interpretation \cite{TIT_Matsu_91, CTW_Matsu_94, CTW_Matsu_07,  kontoyiannis_22} of the context tree weighting (CTW) algorithm \cite{CTW95} that has been originally studied in the field of data compression in information theory. The BCT models have been studied by Matsushima et al. in the field of data compression\cite{CTW_Matsu_94, CTW_Matsu_07, CTW_Matsu_09, Quadtree21, Quadtree22, ChangingContextTree}, their theoretical analysis\cite{Goto_98, Goto_01, Miya_14, Saito_15, Saito_15_IEICE, Saito_16}, and applications to various other fields \cite{full_rooted_trees, rooted_trees, NakaharaISIT23, dobashi_entropy21, Nakahara_batch_metatree, ichijo_isita24, nakahara_aistats25, ichijo_mlsp25, TreeGMM, TSSBGMM, Nakahara_ICSDS, Wang_ISITA, SoftBCT_arXiv}. Recently, the BCT models have been revisited by Kontoyiannis et al. and applied to even more fields \cite{PK21, kontoyiannis_22, PK_22, kontoyiannis_cange_point, PK_23_ITW, PK_24, PK25}. For historical background of the BCT models, see \cite{SoftBCT_arXiv}. The FSBT model \cite{Nakahara_ICSDS} is one of these applications. In the FSBT model, recursive partitioning of the time indices is represented by a tree, and a special prior distribution \cite{full_rooted_trees} is assumed for this tree. Time series segmentation is performed by estimating its posterior distribution by using an algorithm similar to CTW.

However, the FSBT model has a limitation, i.e., the candidates for recursive split points are fixed at the midpoint of each interval. Therefore, an unnecessarily deep tree is often required to represent the segmentation (see also Fig.\ \ref{fig:fixed_splitting} in Section \ref{NumericalExpreiments}). In this study, we propose a variable splitting binary tree (VSBT) model in which interval partitioning is represented by recursive logistic regression models. In this model, split positions other than the midpoint of each interval can be represented by changing the logistic regression coefficients (see also Fig.\ \ref{fig:variable_splitting} in Section \ref{NumericalExpreiments}). This enables more compact tree representation.

On the other hand, it is necessary to simultaneously estimate the logistic regression coefficients (i.e., the split positions) and the tree depth (i.e., the number of splits) to estimate such a model from data. Therefore, we propose an effective estimation algorithm that combines the local variational approximation for logistic regression \cite{Jaakkola-Jordan} with the CTW. We also present numerical examples on synthetic data demonstrating the effectiveness of this algorithm.

Finally, we clarify the difference between previous studies \cite{ChangingContextTree, kontoyiannis_cange_point} and this study. Although these previous studies apply the BCT model to time series with change points, the tree represents partitioning of the context space, i.e., the data value domain, and the probabilistic data generative model within each interval is represented using this tree. In contrast, in this study, the tree represents partitioning of the time domain, and the change process of the probabilistic data generative model is represented using this tree.

\section{Proposed Model}

Herein, we propose the VSBT model. Section \ref{subsection_data_generative_model} describes a data generative model and Section \ref{subsection_priors} explains prior distributions. The graphical model in Fig.\ \ref{fig:graphical_model} will be useful for understanding our model.

\begin{figure}[t]
    \centering
    \includegraphics[width=0.8\linewidth]{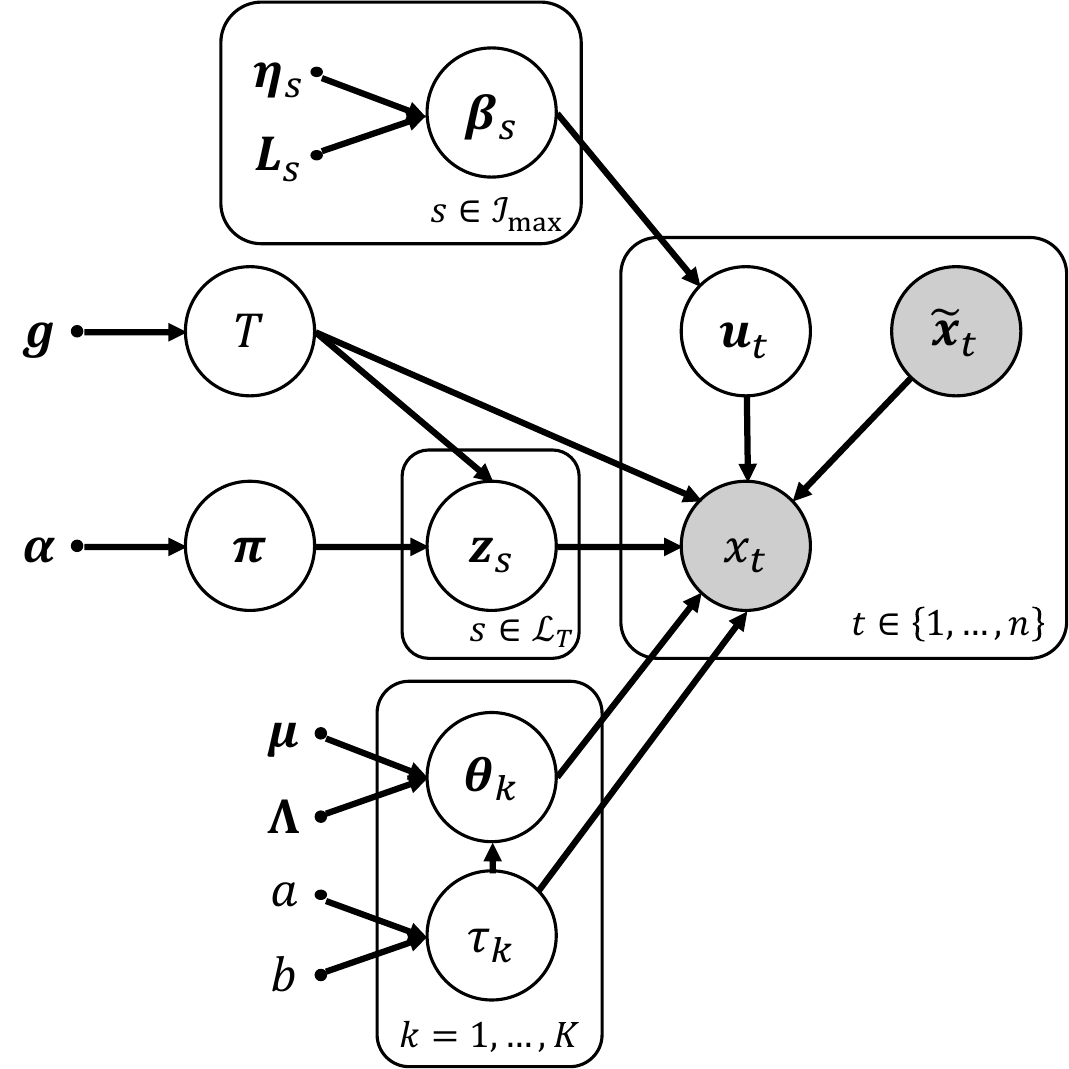}
    \caption{The graphical model of our proposed model. We denote observed variables by shading the corresponding nodes.}
    \label{fig:graphical_model}
\end{figure}

\subsection{Data generative model} \label{subsection_data_generative_model}

\subsubsection{Variable splitting}

Let $T_\mathrm{max}$ be a binary perfect rooted tree with the root node $s_\lambda$ and the depth $D_\mathrm{max}$, where $D_\mathrm{max} \in \mathbb{N}$ is a known constant. For $T_\mathrm{max}$, let $\mathcal{I}_\mathrm{max}$ (resp.\ $\mathcal{L}_\mathrm{max}$) denote the set of all inner nodes (resp.\ leaf nodes) of $T_\mathrm{max}$, and $\mathcal{S}_\mathrm{max} \coloneqq \mathcal{I}_\mathrm{max} \cup \mathcal{L}_\mathrm{max}$. For a node $s \in \mathcal{S}_\mathrm{max}$, $d_s \in \{ 0, 1, \dots , D_\mathrm{max} \}$ denotes the depth of $s$. For a node $s \in \mathcal{I}_\mathrm{max}$, $s_\mathrm{ch}$ denotes a child node of $s$, and more specifically, $s_0$ and $s_1$ denote the left and the right child node of $s$, respectively. The notation $s' \preceq s$ means that $s'$ is an ancestor node of $s$ or $s'$ equals $s$. When we use the notation $s' \prec s$, we exclude the case where $s'$ equals $s$. $\mathrm{Ch}(s)$ denotes the set of all child nodes of $s$.

For $T_\mathrm{max}$, let $\bm u_t = [u_{t,0}, u_{t,1}, \ldots, u_{t,D_\mathrm{max} -1}]^\top \in \{0, 1\}^{D_{\max}}$ be a path vector for time index $t$, where $u_{t,d}=1$ indicates taking the right branch and $u_{t,d}=0$ indicates taking the left branch at depth $d$. The probability distribution of $\bm u \coloneqq \{\bm u_t\}_{t=1}^n$ is defined as 
\begin{align}
    &p(\bm u|\bm \beta) \nonumber \\
    &= \prod_{t=1}^n \prod_{s \in \Imax} \big\{ \sigma(\bm \beta_s^\top \tilde{\bm t})^{u_{t, d_s}} (1 - \sigma(\bm \beta_s^\top \tilde{\bm t}))^{1 - u_{t, d_s}} \big\}^{I\{s \preceq \mathsf{s}(\bm u_t)\}},
\end{align}
where
\begin{itemize}
    \item $\sigma(\cdot)$ denotes the logistic sigmoid function.
    \item $\tilde{\bm t} \coloneqq [t, 1]^\top$.
    \item $I\{\cdot \}$ is the indicator function.
    \item $\mathsf{s}(\bm u_t) \in \Lmax$ is the leaf node determined by $\bm u_t$.
    \item $\bm \beta = \{\bm \beta_s \}_{s \in \Imax}$ is the logistic regression coefficients.
\end{itemize}
We assume a prior distribution $p(\bm \beta)$ (see \eqref{eq_prior_beta}).

\subsubsection{Assignment of autoregressive model}
We define $\mathcal{T}$ as the set of all regular\footnote{All the nodes have either exactly two children or no children.} rooted sub-trees of $T_\mathrm{max}$ whose root node is $s_\lambda$. For $T \in \mathcal{T}$, we assume a prior distribution $p(T)$ (see \eqref{eq_prior_T}). The set of all inner nodes and leaf nodes of $T$ are denoted by $\mathcal{I}_T$ and $\mathcal{L}_T$, respectively.

For each leaf node $s \in \mathcal{L}_T$, we assign one of $K$ candidate autoregressive (AR) models of order $D$.
Let $\bm z_s = [z_{s,1}, z_{s,2}, \ldots, z_{s,K}]^\top \in \{0, 1\}^K$ be a one-hot vector (i.e., one of the elements equals 1 and all remaining elements equal 0) indicating the AR model assigned to the leaf node $s \in \Lset$, i.e., if the $k$-th AR model is assigned to $s \in \Lset$, then $z_{s,k}=1$ and all the remaining elements of $\bm z_s$ equal 0. The assignment $\bm z=\{\bm z_s \}_{s \in \Lset}$ follows a categorical distribution parameterized by $\bm \pi = [\pi_1, \pi_2, \ldots,\pi_K]^\top \in [0, 1]^K$ satisfying $\sum_{k=1}^K \pi_k = 1$:
\begin{align}
    p(\bm z|\bm \pi, T) = \prod_{s \in \Lset} \prod_{k=1}^K \pi_k^{z_{s,k}}, 
\end{align}
where a prior distribution $p(\bm \pi)$ is assumed (see \eqref{eq_prior_pi}).

\subsubsection{Generation of data points}

Given the tree structure $T$, path $\bm u$, and assignment $\bm z$, the probability distribution of a time series $\bm x = (x_1, \dots, x_n)^\top \in \R^n$ is defined as
\begin{align}
    &p(\bm x |\bm u, \bm z, \bm \theta, \bm \tau, T) \nonumber \\
    &= \prod_{t=1}^n \prod_{s \in \Lset} \prod_{k=1}^K \left( \mathcal{N}(x_t | \tilde{\bm x}_t^\top \bm \theta_k, \tau_k^{-1}) \right)^{z_{s,k} I\{s \preceq \mathsf{s}(\bm u_t)\}},
\end{align}
where $\tilde{\bm x}_t = [x_{t-1}, x_{t-2}, \ldots, x_{t-D}, 1]^\top \in \R^{D+1}$ is the vector of past observations, $\bm \theta_k \in \R^{D+1}$ denotes coefficients, and $\tau_k \in \R_{>0}$ denotes precision (inverse of variance). That is, $x_t$ is generated from an AR model assigned to a node $s$ that belongs to $\mathcal{L}_T$ and satisfies $s \preceq \mathsf{s}(\bm u_t)$. Such a node is uniquely determined from $T$ and $\bm u_t$. We assume that the AR parameters $\bm \theta = \{\bm \theta_k \}_{k=1}^{K}$ and $\bm \tau = \{\tau_k\}_{k=1}^{K}$ follow a conjugate Gauss-gamma prior (see \eqref{eq_prior_theta_tau}).

\subsection{Prior distributions} \label{subsection_priors}

We assume a Gaussian prior for $\bm \beta$:
\begin{align}
    p(\bm \beta) = \prod_{s \in \Imax} \mathcal{N}(\bm \beta_s | \bm \eta_s, \bm L_s^{-1}), \label{eq_prior_beta}
\end{align}
where $\bm \eta_s \in \R^2$ and a positive definite matrix $\bm L_s \in \R^{2 \times 2}$ are hyperparameters.

The prior distribution for the tree structure $T$ is governed by splitting probabilities $\bm g = \{g_s\}_{s \in \Smax}$, where $g_s \in [0,1]$ is the probability that node $s$ splits into children. The prior probability of a tree $T$ is given by
\begin{align}
    p(T) = \left( \prod_{s \in \Iset} g_s \right) \left( \prod_{s \in \Lset} (1 - g_s) \right), \label{eq_prior_T}
\end{align}
where $g_s = 0$ for all $s \in \Lmax$.

\begin{Remark}
    The prior in \eqref{eq_prior_T} was introduced by \cite{CTW_Matsu_07}, \cite{CTW_Matsu_09}, and the properties of \eqref{eq_prior_T}, e.g., $\sum_{T \in \in \mathcal{T}} p(T) = 1$, were summarized in \cite{full_rooted_trees}. After these papers, \eqref{eq_prior_T} was revisited by \cite{PK_22}, \cite{PK_24}. 
\end{Remark}

The prior distribution of $\bm \pi$ is the Dirichlet distribution:
\begin{align}
    p(\bm \pi) = \text{Dir}(\bm \pi|\bm \alpha), \label{eq_prior_pi}
\end{align}
where $\bm \alpha \in \R_{>0}^K$ is the hyperparameter.

The prior distribution of $(\bm \theta, \bm \tau)$ is the Gauss-gamma distribution:
\begin{align}
    p(\bm \theta, \bm \tau) = \prod_{k=1}^K \mathcal{N}(\bm \theta_k | \bm \mu, (\tau_k \bm \Lambda)^{-1}) \text{Gam}(\tau_k | a, b), \label{eq_prior_theta_tau}
\end{align}
where $\bm \mu \in \R^{D+1}$, a positive definite matrix $\bm \Lambda \in \R^{(D+1) \times (D+1)}$, $a \in \R_{>0}$, and $b \in \R_{>0}$ are hyperparameters.

\section{Learning Algorithm}

\subsection{Problem setup} 

We consider the problem of segmenting a time series $\bm x = (x_1, \dots, x_n)^\top \in \R^n$ of length $n$. The segmentation is achieved by estimating $\bm z$, $T$, and $\bm \beta$. When we assume the 0-1 loss, the optimal decision based on Bayesian decision theory is given by maximum a posteriori (MAP) estimation (see, e.g., \cite{berger1985statistical}). However, the posterior distribution does not have closed-form expression in our setting. Then, we use the technique of variational inference (see e.g., \cite[Chapter 10]{bishop}).

\subsection{Variational inference with local variational approximation} 

In variational inference, we approximate the joint posterior distribution $p(\bm u, \bm z, T, \bm \theta, \bm \tau, \bm \pi, \bm \beta | \bm x)$ by a variational distribution $q(\bm u, \bm z, T, \bm \theta, \bm \tau, \bm \pi, \bm \beta)$. Hereafter, we assume the following factorization property as a condition: $q(\bm u, \bm z, T, \bm \theta, \bm \tau, \bm \pi, \bm \beta) = q(\bm u) q(\bm z, T) q(\bm \theta, \bm \tau, \bm \pi, \bm \beta)$.

It is known (e.g., \cite{bishop}) that minimizing the Kullback-Leibler divergence between $q(\bm u, \bm z, T, \bm \theta, \bm \tau, \bm \pi, \bm \beta)$ and $p( \bm u, \bm z, T, \bm \theta, \bm \tau, \bm \pi, \bm \beta | \bm x )$ is equivalent to maximizing the variational lower bound $\mathrm{VL}(q)$, which is defined as follows:
\begin{align}
    \mathrm{VL}(q) \coloneqq \mathbb{E}_{q(\bm u, \bm z, T, \bm \theta, \bm \tau, \bm \pi, \bm \beta)} \left[ \ln \frac{p(\bm x, \bm u, \bm z, T, \bm \theta, \bm \tau, \bm \pi, \bm \beta)}{q(\bm u, \bm z, T, \bm \theta, \bm \tau, \bm \pi, \bm \beta)}\right].
\end{align}
However, the optimal variational distribution that maximizes $\mathrm{VL}(q)$ has not closed-form expression yet. It is because our model contains the logistic regression models. Then, we derive a lower bound of $\mathrm{VL}(q)$ and maximize it. Such a method is called local variational approximation\cite{Jaakkola-Jordan}.

We introduce an additional parameter $\xi_{s,t}$ for each inner node $s$ and time $t$. By using the bound proposed in \cite{Jaakkola-Jordan}, the following holds for any $\xi_{s,t} \in \mathbb{R}$ and each logistic regression model.
\begin{align}
    &\sigma(\bm \beta_s^\top \tilde{\bm t})^{u_{t, d_s}} (1 - \sigma(\bm \beta_s^\top \tilde{\bm t}))^{1 - u_{t, d_s}} \nonumber \\
    &\geq \sigma (\xi_{s,t}) \exp \bigl\{ \bm \beta_s^\top \tilde{\bm t} u_{t,d_s} -\frac{1}{2} (\bm \beta_s^\top \tilde{\bm t} + \xi_{s,t}) \nonumber \\
    &\qquad \qquad \qquad \qquad \quad - \lambda(\xi_{s,t}) ((\bm \beta_s^\top \tilde{\bm t})^2 - \xi_{s,t}^2 ) \bigr\} \nonumber \\
    &\eqqcolon h(u_{t,d_s} | \bm \beta_s, \xi_{s,t}),
\end{align}
where $\lambda (\xi_{s,t}) \coloneqq \frac{1}{2 \xi_{s,t}} (\sigma (\xi_{s,t}) - \frac{1}{2})$. Therefore, $p(\bm u | \bm \beta)$ has the following lower bound:
\begin{align}
    p(\bm u | \bm \beta) & \geq \prod_{t=1}^{n} \prod_{s \in \mathcal{I}_\mathrm{max}} h(u_{t,d_s} | \bm \beta_s, \xi_{s,t})^{I \{ s \preceq \mathsf{s}(\bm u_t) \}} \\
    &\eqqcolon h(\bm u | \bm \beta, \bm \xi), \label{eq_def_h(u)}
\end{align}
where $\bm \xi \coloneqq \{ \xi_{s,t} \}_{s \in \mathcal{I}_\mathrm{max}, t \in \{ 1, \dots , n \}}$. By using this, $\mathrm{VL}(q)$ is lower bounded as follows:
\begin{align}
    &\mathrm{VL}(q) \geq \nonumber \\
    &\mathbb{E}_{q(\bm u, \bm z, T, \bm \theta, \bm \tau, \bm \pi, \bm \beta) } \left[ \ln \frac{p(\bm x, \bm z, T, \bm \theta, \bm \tau, \bm \pi | \bm u) h(\bm u | \bm \beta, \bm \xi) p(\bm \beta)}{q(\bm u, \bm z, T, \bm \theta, \bm \tau, \bm \pi, \bm \beta)}\right]. 
\end{align}

It is known (e.g., \cite{bishop}) that the optimal variational distribution maximizing the above lower bound satisfies
\begin{align}
    &\ln q^* (\bm u) = \mathbb{E}_{q^*(\bm z, T, \bm \theta, \bm \tau, \bm \pi, \bm \beta)} \bigl[ \nonumber \\
    &\quad \ln \bigl( p(\bm x, \bm z, T, \bm \theta, \bm \tau, \bm \pi | \bm u) h(\bm u | \bm \beta, \bm \xi) p(\bm \beta) \bigr) \bigr] + \mathrm{const.}, \label{q_star_u} \\
    &\ln q^* (\bm z, T) = \mathbb{E}_{q^*(\bm u, \bm \theta, \bm \tau, \bm \pi, \bm \beta)} \bigl[ \nonumber \\
    &\quad \ln \bigl( p(\bm x, \bm z, T, \bm \theta, \bm \tau, \bm \pi | \bm u) h(\bm u | \bm \beta, \bm \xi) p(\bm \beta) \bigr) \bigr] + \mathrm{const.}, \label{q_star_z_T} \\
    &\ln q^* (\bm \theta, \bm \tau , \bm \pi, \bm \beta) = \mathbb{E}_{q^*(\bm u, \bm z, T)} \bigl[ \nonumber \\
    &\quad \ln \bigl( p(\bm x, \bm z, T, \bm \theta, \bm \tau, \bm \pi | \bm u) h(\bm u | \bm \beta, \bm \xi) p(\bm \beta) \bigr) \bigr] + \mathrm{const.} \label{q_star_theta_tau_pi_beta}
\end{align}

However, $q^* (\bm u)$, $q^* (\bm z, T)$, and $q^* (\bm \theta, \bm \tau, \bm \pi, \bm \beta)$ depend on each other. Therefore, we update them in turn from an initial value until convergence. These expectations have closed-form expressions, and we show them in the following sections. For details of the derivation, see Appendix. Hereafter, prime symbols (') will be appended to the parameters of the variational distribution to distinguish them from those of the prior distribution.

\subsection{Update formula of \texorpdfstring{\eqref{q_star_u}}{(13)}}

Calculating \eqref{q_star_u}, we can derive a factorization $q(\bm u)=\prod_{t=1}^{n} q(\bm u_t)$, and each $q(\bm u_t)$ is given as
\begin{align}
    q(\bm u_t) = \prod_{s \in \Imax} \prod_{u \in \{0,1\}} (\varpi'_{t,s,s_u})^{I\{s_u \preceq \mathsf{s}(\bm u_t)\}}, 
\end{align}
where $s_0$ and $s_1$ denote the left and right children of $s$, respectively, $\varpi'_{t,s,s_u} \coloneqq \varrho_{t, s, s_u}/\sum_{u \in \{0,1\}} \varrho_{t, s, s_u}$,
and $\varrho_{t, s, s_u} $ is given as 
\begin{align}
    &\ln \varrho_{t, s, s_u} \coloneqq \nonumber \\
    & \begin{cases}
        (\clubsuit) + (\diamondsuit) + \ln \sum_{s_\mathrm{ch} \in \mathrm{Ch}(s_u)} \varrho_{t,s_u,s_\mathrm{ch}}, & s_u \in \mathcal{I}_\mathrm{max}, \\
        (\clubsuit) + (\diamondsuit), & s_u \in \mathcal{L}_\mathrm{max}.
    \end{cases}
\end{align}
Here, $(\clubsuit)$ is 
\begin{align}
    (\clubsuit) = & \mathbb{E}_{q(\bm \beta_s)}[ \ln h( u | \bm \beta_s, \xi_{s,t})] \\
    = &\ln \sigma(\xi_{s,t}) + (\bm \eta'_s)^\top \tilde{\bm t} u  -\frac{1}{2} \left((\bm \eta'_s)^\top \tilde{\bm t} + \xi_{s,t} \right) \nonumber \\
    & - \lambda(\xi_{s,t}) \left(\tilde{\bm t}^\top ( \bm \eta'_s (\bm \eta'_s)^\top + (\bm L'_s)^{-1} )\tilde{\bm t} -  (\xi_{s,t})^2 \right), 
\end{align}
where $\bm L'_s$, $\bm \eta'_s$, and $\xi_{s,t}$ are given as in \eqref{eq_update_L}, \eqref{eq_update_eta}, and \eqref{eq_update_xi}, respectively, and $(\diamondsuit)$ is 
\begin{align}
    & (\diamondsuit) = \mathbb{E}_{q(\bm z, T, \bm \theta, \bm \tau)} \Biggl[ I \{ s_u \in \mathcal{L}_T \} \nonumber \\
    &\qquad \qquad \qquad \quad \times \sum_{k=1}^K z_{s_u, k} \ln \mathcal{N}(x_t | \tilde{\bm x}_t^\top \bm \theta_k, \tau_k^{-1})\Biggr] \\
    &=\frac{1}{2} (1-g'_{s_u}) \left( \prod_{s \prec s_u} g'_{s} \right) \sum_{k=1}^K \pi'_{s_u, k} \bigg\{ \psi(a'_{k}) - \ln b'_{k} \nonumber \\
    &\quad -\ln 2\pi - \frac{a'_{k}}{b'_{k}}(x_t - (\tilde{\bm x}_t)^\top \bm \mu'_{k})^2 - (\tilde{\bm x}_t)^\top (\bm \Lambda'_{k})^{-1} \tilde{\bm x}_t \bigg\},
\end{align}
where $\psi(\cdot)$ denotes the digamma function, and $\pi'_{s, k}$ $g'_s$, $\bm \Lambda'_{k}$, $\bm \mu'_{k}$, $a'_{k}$, and $b'_{k}$ are given as in \eqref{eq_update_pi}, \eqref{eq_update_g}, \eqref{eq_update_lambda}, \eqref{eq_update_mu}, \eqref{eq_update_a}, and \eqref{eq_update_b}, respectively.

\subsection{Update formula of \texorpdfstring{\eqref{q_star_z_T}}{(14)}}

Calculating \eqref{q_star_z_T}, we can derive a factorization $q(\bm z, T) = q(T) \prod_{s \in \Lset}$ $q(\bm z_s)$. For each $s \in \Lset$, $q(\bm z_s)$ is given as
\begin{align}
    q(\bm z_s) = \prod_{k=1}^{K} (\pi'_{s,k})^{z_{s,k}},\text{ where } \pi'_{s,k} \coloneqq \frac{\rho_{s,k}}{\sum_{k=1}^{K} \rho_{s,k}} \label{eq_update_pi}
\end{align}
and $\rho_{s,k}$ is given as
\begin{align}
    \ln \rho_{s,k} \coloneqq& \mathbb{E}_{q(\bm \pi)}[ \ln \pi_k] \nonumber \\
    &\quad + \sum_{t=1}^{n} q_{s,t} \mathbb{E}_{q(\bm \theta, \bm \tau)} [\ln \mathcal{N}(x_t | \tilde{\bm x}_t^\top \bm \theta_k, \tau_k^{-1})] \\
    =& \psi(\alpha'_k) - \psi \left( \sum_{k=1}^{K} \alpha'_k \right) \nonumber \\
    &\quad - \frac{1}{2}\mathrm{Tr} \{ \bm Q_s \} (\ln 2\pi - \psi(a'_{k}) + \ln b'_{k}) \nonumber \\
    &\quad - \frac{a'_{k}}{2 b'_{k}}(\bm x - \bm X \bm \mu'_{k})^\top  \bm Q_s (\bm x - \bm X \bm \mu'_{k}) \nonumber \\
    &\quad - \frac{1}{2} \mathrm{Tr} \{ \bm X^\top \bm Q_s \bm X (\bm \Lambda'_k)^{-1} \},
\end{align}
where
\begin{itemize}
    \item $q_{s,t}$ is the posterior probability that $s \preceq \mathsf{s}(\bm u_t)$ holds, i.e., $q_{s,t} \coloneqq \prod_{s', s'_\mathrm{ch} \preceq s} \varpi'_{t, s', s'_\mathrm{ch}}$.
    \item $\bm Q_s \in \mathbb{R}^{n \times n}$ is the diagonal matrix defined as $\mathrm{diag} \left \{ q_{s,1}, q_{s,2}, \dots , q_{s,n} \right \}$.
    \item $\bm X \in \mathbb{R}^{n \times (D+1)}$ denotes the matrix whose $t$-th row is $\tilde{\bm x}_t^\top$.
    \item $\mathrm{Tr} \{\cdot\}$ denotes the trace of a matrix.
    \item $\alpha'_k$ is given as in \eqref{eq_update_alpha}.
\end{itemize}

Moreover, the update formula for $q(T)$ is given as follows:
\begin{align}
    q(T) = \left( \prod_{s \in \mathcal{I}_T} g'_s \right) \left( \prod_{s \in \mathcal{L}_T} (1-g'_s) \right), \label{eq_update_formula_Tree}
\end{align}
where
\begin{align}
    g'_s &\coloneqq \! \begin{cases}
        \frac{g_s \prod_{s_\mathrm{ch} \in \mathrm{Ch}(s)} \phi_{s_\mathrm{ch}}}{\phi_s}, & s \in \mathcal{I}_\mathrm{max}, \\
        0, & s \in \mathcal{L}_\mathrm{max},
    \end{cases} \label{eq_update_g} \\
    \phi_s &\coloneqq \! \begin{cases}
        (1-g_s)\sum_{k=1}^K \rho_{s,k} + g_s \prod_{s_{\mathrm{ch}} \in \mathrm{Ch}(s)} \phi_{s_{\mathrm{ch}}}, & \! \! s \in \Imax, \\
        \sum_{k=1}^K \rho_{s,k}, & \! \! s \in \Lmax.
    \end{cases} \label{eq_update_formula_phi}
\end{align}

\begin{Remark}
    The update formula \eqref{eq_update_formula_phi} has a similar structure to the CTW algorithm \cite{CTW95} (to be more precise, the Bayes coding algorithm for context tree models \cite{CTW_Matsu_07}, \cite{CTW_Matsu_09}). 
\end{Remark}

\subsection{Update formula of \texorpdfstring{\eqref{q_star_theta_tau_pi_beta}}{(15)}}

Calculating \eqref{q_star_theta_tau_pi_beta}, we can derive a factorization $q(\bm \theta, \bm \tau, \bm \pi,$ $\bm \beta) = q(\bm \pi) \prod_{k=1}^{K}q(\bm \theta_k, \bm \tau_k) \prod_{s \in \mathcal{I}_\mathrm{max}} q(\bm \beta_s)$. First, the posterior $q(\bm \pi)$ is given as $q(\bm \pi) = \mathrm{Dir}(\bm \pi | \bm \alpha')$, where each element of $\bm \alpha'$ is given as
\begin{align}
    \alpha'_k \coloneqq \alpha_k + \sum_{s \in \Smax} (1-g'_{s}) \left(\prod_{s' \prec s} g'_{s'} \right) \pi'_{s,k}. \label{eq_update_alpha}
\end{align}

Second, the posterior $q(\bm \theta_k, \bm \tau_k)$ is given as $q(\bm \theta_k, \bm \tau_k) = \mathcal{N}(\bm \theta_k | \bm \mu'_k, (\tau_k \bm \Lambda'_k)^{-1}) \text{Gam}(\tau_k | a'_k, b'_k)$, where 
\begin{align}
    &\bm \Lambda'_k  \coloneqq  \bm \Lambda + \sum_{s \in \Smax} (1-g'_{s}) \left(\prod_{s' \prec s} g'_{s'} \right)\pi'_{s,k} \bm X^\top \bm Q_s \bm X, \label{eq_update_lambda} \\
    &\bm \mu'_k  \coloneqq  \bigl( \bm \Lambda'_k \bigr)^{-1} \Biggl( \bm \Lambda \bm \mu \nonumber \\
    & \qquad \quad+ \sum_{s \in \Smax} (1-g'_{s}) \left(\prod_{s' \prec s} g'_{s'} \right)\pi'_{s,k} \bm X^\top \bm Q_s \bm x \Biggr), \label{eq_update_mu} \\
    & a'_k  \coloneqq   a  +  \frac{1}{2} \sum_{s \in \Smax} (1 - g'_{s}) \left(\prod_{s' \prec s} g'_{s'} \right)\pi'_{s,k} \mathrm{Tr} \{ \bm Q_s \}, \label{eq_update_a}\\
    & b'_k  \coloneqq  b  +  \frac{1}{2} \Biggl( \bm \mu^\top \bm \Lambda \bm \mu - (\bm \mu'_k)^\top \bm \Lambda'_k \bm \mu'_k \nonumber \\
    &\qquad \quad + \sum_{s \in \Smax} (1-g'_{s}) \left(\prod_{s' \prec s} g'_{s'} \right)\pi'_{s,k} \bm x^\top \bm Q_s \bm x \Biggr). \label{eq_update_b}
\end{align}

Lastly, the posterior $q(\bm \beta_s)$ is given as $q(\bm \beta_s) = \mathcal{N}(\bm \beta_s | \bm \eta'_s, (\bm L'_s)^{-1})$, where $\bm \eta'_s$ and $\bm L'_s$ are given as
\begin{align}
    \bm L'_s &\coloneqq \bm L_s + 2 \sum_{t=1}^n q_{s,t} \lambda(\xi_{s,t}) \tilde{\bm t} \tilde{\bm t}^\top, \label{eq_update_L} \\
    \bm \eta'_s &\coloneqq (\bm L'_s)^{-1} \left( \bm L_s \bm \eta_s + \sum_{t=1}^n q_{s,t} \left( \varpi'_{t,s,s_1} - \frac{1}{2} \right) \tilde{\bm t} \right). \label{eq_update_eta}
\end{align}
The parameter $\xi_{s,t}$ is updated as 
\begin{align}
    \xi_{s,t} \coloneqq \left( \tilde{\bm t}^\top ( (\bm L'_s)^{-1} + \bm \eta'_s ({\bm \eta'_s})^\top ) \tilde{\bm t} \right)^{1/2}. \label{eq_update_xi}
\end{align}

\subsection{Initialization}

Since the variational inference is a greedy algorithm, the initialization method is crucial. Our algorithm is also highly sensitive to the initial value settings. Here, as an example, we describe the initialization method used in the subsequent numerical experiments.

When using this method, we assume $K = 2^{D_\mathrm{max}}$. First, we search for split points using a decision tree construction algorithm that employs the marginal likelihood as an evaluation criterion. More specifically, let $\bm x_1$ denote the data points belonging to the left interval after splitting and $\bm x_2$ denote those belonging to the right interval. We greedily search for split points from the root node to the nodes at maximum depth so as to maximize the sum of log marginal likelihoods $\sum_{i = 1}^2 \ln \int p(\bm x_i | \bm \theta, \bm \tau) p(\bm \theta, \bm \tau) \mathrm{d}\bm \theta \mathrm{d}\bm \tau$, assuming each of them is generated from a single AR model. Next, we compute $q(\bm u)$ under the assumption that the obtained thresholds are deterministic change points. (That is, $\bm \varpi'_{t,s}$ becomes a one-hot vector for any $t$ and $s$.)

After that, we repeat only the update of $q(\bm \beta)$ until convergence. The initial value of $q(\bm \beta)$ here is set as $\bm \eta'_s = [1.0, -h_s]^\top$, where $h_s$ is the split point of node $s$ obtained by the search.

Finally, we determine $q(\bm z, T)$ as follows. For any $s \in \mathcal{I}_\mathrm{max}$, we set $g'_s=1.0$ and $\bm \pi'_s = [1/K, \dots , 1/K]^\top$. For any $s \in \mathcal{L}_\mathrm{max}$, if $s$ is the $j$-th node from the left ($j = 1, 2, \dots , 2^{D_\mathrm{max}}$), we set $g'_s = 0.0$ and $\pi'_{s,k} = I \{ k=j \}$. Then, we begin the iterative updates starting from $q(\bm \theta, \bm \tau, \bm \pi)$.

\section{Numerical Experiments}\label{NumericalExpreiments}

\subsection{Experiment 1: effect of variable splitting}

In Experiment 1, we verify the effect of variable splitting. We use synthetic data generated as follows. From time 1 to 25 and 51 to 75, the data are generated from a first-order AR model whose coefficients and variance are $[0.8, 2.0]^\top$ and $1.0$, respectively. From time 26 to 50, they are generated from another first-order AR model whose coefficients and variance are $[0.8, -2.0]^\top$ and $1.0$, respectively. 

The hyperparameters are set as follows. For $\bm \eta_s$, when node $s$ is the $j$-th node from the left at depth $d$, we set $\bm \eta_s^\top = [\eta_{s,1}, \eta_{s,2}] = [1.0, -2^{-(d+1)}(2j-1)n]$. In other words, letting $h_s$ be the split point such that $\eta_{s,1} h_s + \eta_{s,2} = 0$, we determine $\bm \eta_s$ so that recursive splitting is performed at the midpoint of each interval. For hyperparameters other than $\bm \eta_s$, we use the following values: $D=1$, $D_\mathrm{max} = 5$, $K = 2^{D_\mathrm{max}} = 32$, $\bm L_s = \bm I$ and $g_s=0.5$ for any $s$, $\bm \alpha = [1/2, \dots , 1/2]^\top$, $\bm \mu = \bm 0$, $\bm \Lambda = \bm I$, $a = 1.0$, and $b = 1.0$. We use the initialization method described in the previous section.

\begin{figure}[t]
    \centering
    \includegraphics[width=1.0\linewidth]{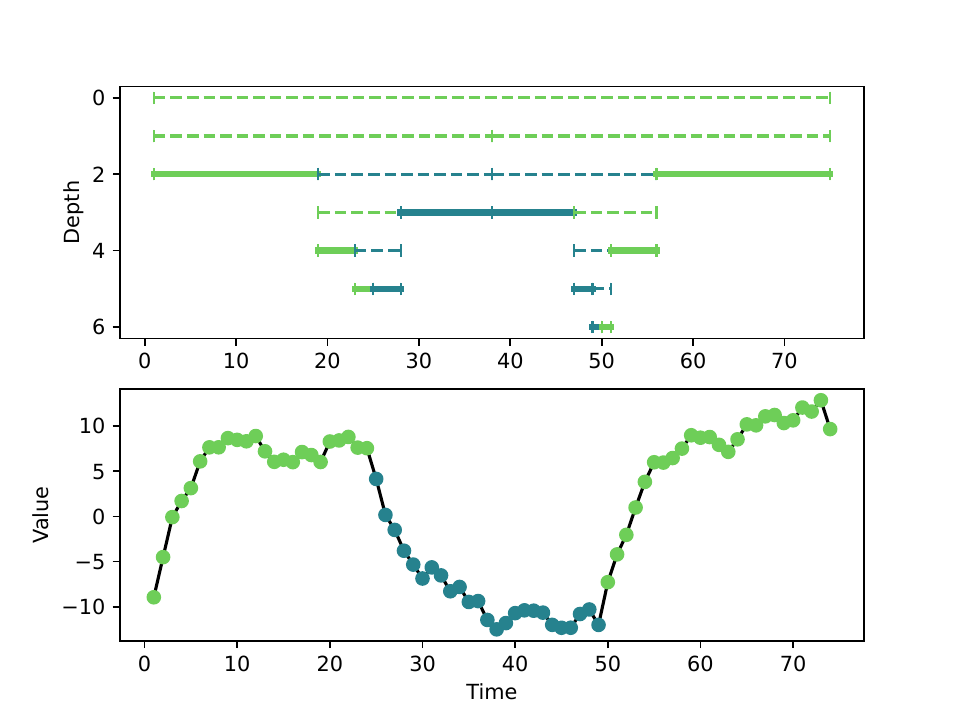}
    \caption{The segmentation under the FSBT model}
    \label{fig:fixed_splitting}
\end{figure}

\begin{figure}[t]
    \centering
    \includegraphics[width=1.0\linewidth]{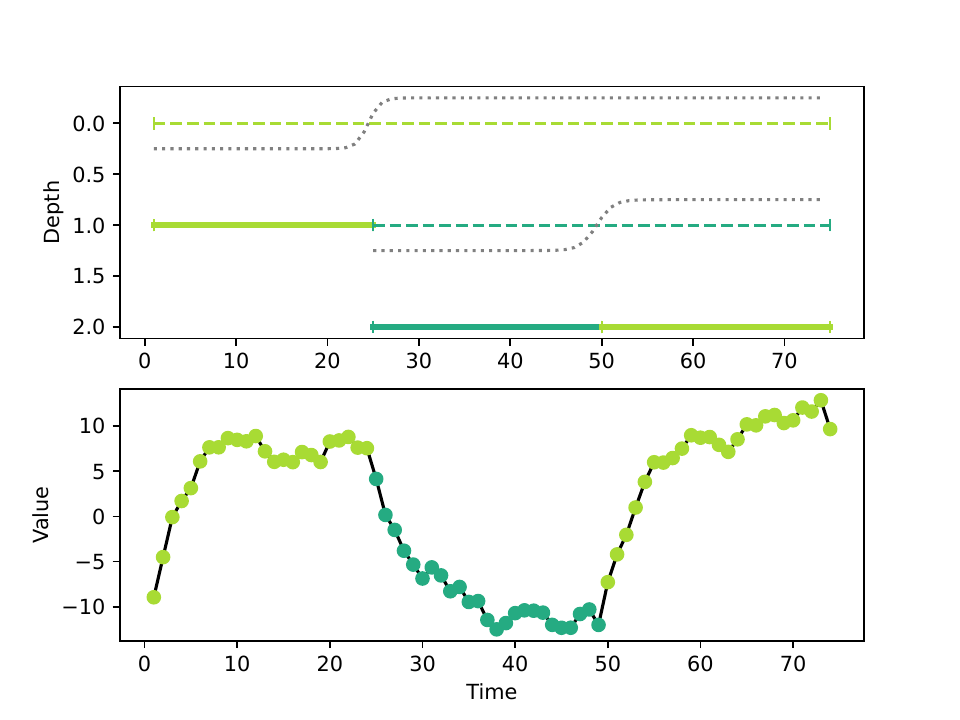}
    \caption{The segmentation under the VSBT model}
    \label{fig:variable_splitting}
\end{figure}

Figures \ref{fig:fixed_splitting} and \ref{fig:variable_splitting} show the estimated models under the conventional FSBT and the VSBT model obtained by using a MAP tree estimation algorithm similar to \cite{full_rooted_trees}, respectively. For the FSBT model, the maximum depth was set to $D_\mathrm{max}=10$. The upper part of each figure shows the estimated tree structure and split points, while the lower part shows the data used for training. The colors of the intervals and plots represent the indices of the estimated AR models.

Although the FSBT model is able to find the change points, it requires a deep tree to represent the segmentation. On the other hand, we can see that the proposed VSBT model can represent the segmentation with a tree of minimal depth.

\subsection{Experiment 2: uncertainty quantification of changes}

In Experiment 2, we evaluate the uncertainty of changes at each time point, which is an advantage of Bayesian methods. We use another synthetic data created by adding a noise to a sine wave. The order of the AR model in our model is set to $D=0$, that is, we assume that the data points follow i.i.d. normal distributions. The other hyperparameters, including the initialization method, are the same as in Experiment 1.

The posterior probability of the change at each time point is calculated as follows. First, let $\bm \zeta_t$ be a one-hot vector such that $\zeta_{t, k} = 1$ holds when the AR model at time $t$ is the $k$-th model. If $\bm \zeta_t \neq \bm \zeta_{t+1}$, we say that there was a change in the model between time $t$ and $t+1$. Then, we calculate the posterior probability of that change as follows. First, let $\bm r_t$ be a vector whose elements represent the probabilities that $\zeta_{t,k} = 1$ holds under the posterior distribution. The vector $\bm r_t$ can be calculated as $\bm r_t = \bm r_{t, s_\lambda}$ by using $\bm r_{t, s}$ defined below.
\begin{align}
    \bm r_{t,s} \coloneqq \begin{cases}
        \bm \pi'_s, & \! s \in \mathcal{L}_\mathrm{max}, \\
        (1-g'_s) \bm \pi'_s + g'_s \sum_{u=\{ 0, 1 \}} \varpi'_{t,s,s_u} \bm r_{t,s_u}, & \! s \in \mathcal{I}_\mathrm{max}.
    \end{cases}
\end{align}
Then, we calculate the posterior probability of the change as $1 - \bm r_t^\top \bm r_{t+1}$. (Strictly speaking, since $\bm \zeta_t$ and $\bm \zeta_{t+1}$ are not independent under the posterior distribution, this calculation method involves an approximation.)

\begin{figure}[t]
    \centering
    \includegraphics[width=1.0\linewidth]{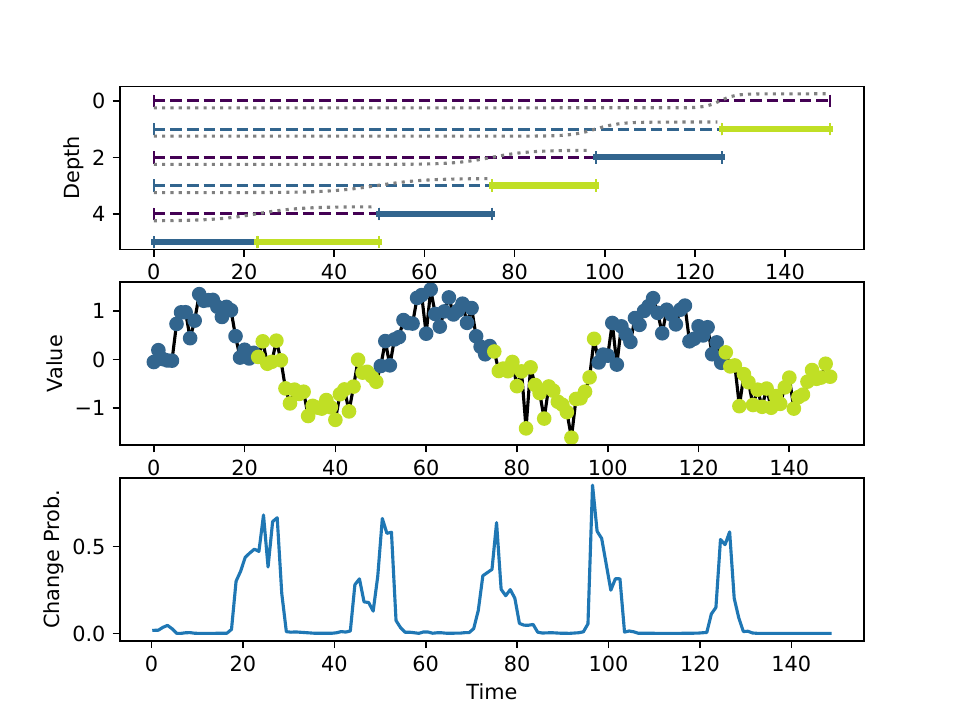}
    \caption{The segmentation and the posterior probabilities of the changes}
    \label{fig:sin_segmentation_change_prob}
\end{figure}

Figure \ref{fig:sin_segmentation_change_prob} shows the estimated model and the posterior probability of the changes in Experiment 2. We can see that the posterior probability of the changes varies smoothly around the split points of the MAP-estimated model. This enables us to evaluate the uncertainty of the changes, which cannot be obtained from the point estimation of the model alone.

\section{Conclusion}

We proposed a variable splitting binary tree (VSBT) model based on the BCT models for time series segmentation. The VSBT model can represent split positions at arbitrary locations within each interval by recursive logistic regression models. To simultaneously estimate both the logistic regression coefficients (i.e., the split positions) and the tree depth (i.e., the number of splits), we proposed an effective algorithm that combines the local variational approximation for logistic regression with the CTW.

\section*{Acknowledgment}
This work was supported in part by JSPS KAKENHI Grant Numbers JP23K03863, JP23K04293, JP23H00468, JP24H00370, JP25K07732, and JST SPRING Grant Number JPMJSP2128.

\bibliographystyle{IEEEtran}
\bibliography{refs}

\appendix

The following lemmas, Lemma \ref{lemma_q_u}, Lemma \ref{lemma_q_z_T} and Lemma \ref{lemma_q_theta_tau_pi_beta}, give the form of the posterior distributions $q(\bm u)$, $q(\bm z, T)$, and $q(\bm \theta, \bm \tau, \bm \pi, \bm \beta)$, respectively. Note that the calculation of the expectations included in these lemmas will be summarized later in Lemma \ref{lemma_expectations}.

\begin{Lemma}\label{lemma_q_u}
The posterior $q(\bm u)$ can be factorized as $\prod_{t=1}^{n} q(\bm u_t)$, and each $q(\bm u_t)$ is given as 
\begin{align}
    q(\bm u_t) = \prod_{s \in \mathcal{I}_\mathrm{max}} \prod_{u \in \{ 0, 1 \}} (\varpi'_{t,s,s_u})^{I \{ s_u \preceq \mathsf{s}(\bm u_t) \}}, \label{eq:q_u}
\end{align}
where 
\begin{align}
    \varpi'_{t, s, s_u} \coloneqq \frac{\varrho_{t, s, s_u}}{\sum_{u \in \{ 0, 1 \}} \varrho_{t, s, s_u}} \label{eq:varpi}
\end{align}
and
\begin{align}
    &\ln \varrho_{t, s, s_u} \nonumber \\
    &\coloneqq \begin{cases}
        \mathbb{E}_{q(\bm \beta_s)}[ \ln h( u | \bm \beta_s, \xi_{s,t})] \\
        \quad + \mathbb{E}_{q(\bm z, T, \bm \theta, \bm \tau)} [I \{ s_u \in \mathcal{L}_T \} \\
        \qquad \times \sum_{k=1}^K z_{s_u, k} \ln \mathcal{N}(x_t | \tilde{\bm x}_t^\top \bm \theta_k, \tau_k^{-1})] \\
        \quad + \ln \sum_{s_\mathrm{ch} \in \mathrm{Ch}(s_u)} \varrho_{t,s_u,s_\mathrm{ch}}, & s_u \in \mathcal{I}_\mathrm{max}, \\
        \mathbb{E}_{q(\bm \beta_s)}[ \ln h( u | \bm \beta_s, \xi_{s,t})] \\
        \quad + \mathbb{E}_{q(\bm z, T, \bm \theta, \bm \tau)} [I \{ s_u \in \mathcal{L}_T \} \\
        \qquad \times \sum_{k=1}^K z_{s_u, k} \ln \mathcal{N}(x_t | \tilde{\bm x}_t^\top \bm \theta_k, \tau_k^{-1})], & s_u \in \mathcal{L}_\mathrm{max}. \\
    \end{cases} \label{eq:varrho}
\end{align}
\end{Lemma}

\begin{IEEEproof}
Calculating the right-hand side of \eqref{q_star_u}, we obtain
\begin{align}
    &\ln q(\bm u) \nonumber \\
    &= \mathbb{E}_{q(\bm \beta)} [\ln h(\bm u | \bm \beta, \bm \xi)] + \mathbb{E}_{q(\bm z, T, \bm \theta, \bm \tau)} [\ln p(\bm x | \bm u, \bm z, \bm \theta, \bm \tau, T)] \nonumber \\
    &\quad + \mathrm{const.} \\
    &= \sum_{t=1}^n \Biggl\{ \sum_{s \in \mathcal{I}_\mathrm{max}} I \{ s \preceq \mathsf{s}(\bm u_t) \} \mathbb{E}_{q(\bm \beta_s)} [\ln h(u_{t,d_s} | \bm \beta_s, \xi_{s,t})] \nonumber \\
    &\qquad \qquad + \mathbb{E}_{q(\bm z, T, \bm \theta, \bm \tau)} \Biggl[ \sum_{s \in \mathcal{L}_T} I \{ s \preceq \mathsf{s}(\bm u_t) \} \nonumber \\
    & \qquad \qquad \qquad \times \sum_{k=1}^K z_{s,k} \ln \mathcal{N}(x_t | \tilde{\bm x}_t^\top \bm \theta_k, \tau_k^{-1}) \Biggr] \Biggr\} + \mathrm{const.}
\end{align}
Therefore, $q(\bm u)$ can be factorized as $\prod_{t=1}^{n} q(\bm u_t)$. Moreover, $q(\bm u_t)$ is calculated as
\begin{align}
    &\ln q(\bm u_t) \nonumber \\
    &= \sum_{s \in \mathcal{I}_\mathrm{max}} I \{ s \preceq \mathsf{s}(\bm u_t) \} \mathbb{E}_{q(\bm \beta_s)} [\ln h(u_{t,d_s} | \bm \beta_s, \xi_{s,t})] \nonumber \\
    &\quad + \sum_{s \in \mathcal{S}_\mathrm{max}} I \{ s \preceq \mathsf{s}(\bm u_t) \} \mathbb{E}_{q(\bm z, T, \bm \theta, \bm \tau)} \Biggl[ I \{ s \in \mathcal{L}_T \} \nonumber \\
    &\qquad \quad \times \sum_{k=1}^K z_{s,k} \ln \mathcal{N}(x_t | \tilde{\bm x}_t^\top \bm \theta_k, \tau_k^{-1}) \Biggr] + \mathrm{const.} \\
    &= \sum_{s \in \mathcal{I}_\mathrm{max}} \sum_{u \in \{ 0, 1 \}} I \{ s_u \preceq \mathsf{s}(\bm u_t) \} \mathbb{E}_{q(\bm \beta_s)} [\ln h(u | \bm \beta_s, \xi_{s,t})] \nonumber \\
    &\quad + \mathbb{E}_{q(\bm z, T, \bm \theta, \bm \tau)} \Biggl[ I \{ s_\lambda \in \mathcal{L}_T \} \nonumber \\
    &\qquad \quad \times \sum_{k=1}^K z_{s_\lambda,k} \ln \mathcal{N}(x_t | \tilde{\bm x}_t^\top \bm \theta_k, \tau_k^{-1}) \Biggr] \nonumber \\
    &\quad + \sum_{s \in \mathcal{I}_\mathrm{max}} \sum_{u \in \{ 0, 1 \}} I \{ s_u \preceq \mathsf{s}(\bm u_t) \} \nonumber \\
    &\qquad \times \mathbb{E}_{q(\bm z, T, \bm \theta, \bm \tau)} \Biggl[ I \{ s_u \in \mathcal{L}_T \} \nonumber \\
    &\qquad \quad \times \sum_{k=1}^K z_{s_u,k} \ln \mathcal{N}(x_t | \tilde{\bm x}_t^\top \bm \theta_k, \tau_k^{-1}) \Biggr] + \mathrm{const.} \\
    &= \sum_{s \in \mathcal{I}_\mathrm{max}} \sum_{u \in \{ 0, 1 \}} I \{ s_u \preceq \mathsf{s}(\bm u_t) \} \Biggl\{ \mathbb{E}_{q(\bm \beta_s)} [\ln h(u | \bm \beta_s, \xi_{s,t})] \nonumber \\
    &\qquad  + \mathbb{E}_{q(\bm z, T, \bm \theta, \bm \tau)} \Biggl[ I \{ s_u \in \mathcal{L}_T \} \nonumber \\
    &\qquad \quad \times \sum_{k=1}^K z_{s_u,k} \ln \mathcal{N}(x_t | \tilde{\bm x}_t^\top \bm \theta_k, \tau_k^{-1}) \Biggr] \Biggr\} + \mathrm{const.} \label{eq:q_u_goal}
\end{align}

Next, we show \eqref{eq:q_u} reduces to \eqref{eq:q_u_goal} by taking the logarithm of \eqref{eq:q_u} and substituting the definitions of $\varpi'_{t,s,s_u}$ and $\varrho_{t,s,s_u}$ into its right-hand side. In the following, we denote the set of nodes of depth $d$ as $\mathcal{S}_d$ and the set of nodes of depth less than $d$ as $\mathcal{S}_{<d}$.
\begin{align}
    &\sum_{s \in \mathcal{I}_\mathrm{max}} \sum_{u \in \{ 0, 1 \}} I \{ s_u \preceq \mathsf{s}(\bm u_t) \}  \ln \varpi'_{t,s,s_u} \nonumber \\
    &= \sum_{s \in \mathcal{S}_{<D_\mathrm{max}-1}} \sum_{u \in \{ 0, 1 \}} I \{ s_u \preceq \mathsf{s}(\bm u_t) \}  \ln \varpi'_{t,s,s_u} \nonumber \\
    &\quad + \sum_{s \in \mathcal{S}_{D_\mathrm{max}-1}} \sum_{u \in \{ 0, 1 \}} I \{ s_u \preceq \mathsf{s}(\bm u_t) \}  \ln \varpi'_{t,s,s_u} \\
    &= \sum_{s \in \mathcal{S}_{<D_\mathrm{max}-1}} \sum_{u \in \{ 0, 1 \}} I \{ s_u \preceq \mathsf{s}(\bm u_t) \} \nonumber \\
    &\qquad \times \Biggl\{ \mathbb{E}_{q(\bm \beta_s)} [\ln h(u | \bm \beta_s, \xi_{s,t})] \nonumber \\
    &\qquad \qquad + \mathbb{E}_{q(\bm z, T, \bm \theta, \bm \tau)} \Biggl[ I \{ s_u \in \mathcal{L}_T) \} \nonumber \\
    &\qquad \qquad \qquad \qquad \qquad \times \sum_{k=1}^K z_{s_u,k} \ln \mathcal{N}(x_t | \tilde{\bm x}_t^\top \bm \theta_k, \tau_k^{-1}) \Biggr] \nonumber \\
    &\qquad \qquad + \ln \sum_{s_\mathrm{ch} \in \mathrm{Ch}(s_u)} \varrho_{t,s_u,s_\mathrm{ch}} - \ln \sum_{u \in \{ 0, 1 \}} \varrho_{t, s, s_u} \Biggr\} \nonumber \\
    &\quad + \sum_{s \in \mathcal{S}_{D_\mathrm{max}-1}} \sum_{u \in \{ 0, 1 \}} I \{ s_u \preceq \mathsf{s}(\bm u_t) \} \nonumber \\
    &\qquad \times \Biggl\{ \mathbb{E}_{q(\bm \beta_s)} [\ln h(u | \bm \beta_s, \xi_{s,t})] \nonumber \\
    &\qquad \qquad + \mathbb{E}_{q(\bm z, T, \bm \theta, \bm \tau)} \Biggl[ I \{ s_u \in \mathcal{L}_T) \} \nonumber \\
    &\qquad \qquad \qquad \qquad \qquad \times \sum_{k=1}^K z_{s_u,k} \ln \mathcal{N}(x_t | \tilde{\bm x}_t^\top \bm \theta_k, \tau_k^{-1}) \Biggr] \nonumber \\
    &\qquad \qquad - \ln \sum_{u \in \{ 0, 1 \}} \varrho_{t, s, s_u} \Biggr\} \\
    &= \sum_{s \in \mathcal{I}_\mathrm{max}} \sum_{u \in \{ 0, 1 \}} I \{ s_u \preceq \mathsf{s}(\bm u_t) \}  \Biggl\{ \mathbb{E}_{q(\bm \beta_s)} [\ln h(u | \bm \beta_s, \xi_{s,t})] \nonumber \\
    &\qquad + \mathbb{E}_{q(\bm z, T, \bm \theta, \bm \tau)} \Biggl[ I \{ s_u \in \mathcal{L}_T) \} \nonumber \\
    &\qquad \qquad \qquad \qquad \times \sum_{k=1}^K z_{s_u,k} \ln \mathcal{N}(x_t | \tilde{\bm x}_t^\top \bm \theta_k, \tau_k^{-1}) \Biggr] \Biggr\} \nonumber \\
    &\quad - \sum_{u \in \{ 0, 1 \}} I \{ (s_\lambda)_u \preceq \mathsf{s}(\bm u_t) \} \ln \sum_{u \in \{ 0, 1 \}} \varrho_{t, s_\lambda, (s_\lambda)_u}, \label{eq:q_u_with_const}
\end{align}
where the last equality follows because $\ln \sum_{u \in \{ 0, 1 \}} \varrho_{t, s, s_u}$ are canceled for $s$ except $s_\lambda$. Moreover, since
\begin{align}
    &\sum_{u \in \{ 0, 1 \}} I \{ (s_\lambda)_u \preceq \mathsf{s}(\bm u_t) \} \ln \sum_{\tilde{u} \in \{ 0, 1 \}} \varrho_{t, s_\lambda, (s_\lambda)_{\tilde{u}}} \nonumber \\
    &\quad= \ln \sum_{\tilde{u} \in \{ 0, 1 \}} \varrho_{t, s_\lambda, (s_\lambda)_{\tilde{u}}},
\end{align}
the last term of \eqref{eq:q_u_with_const} is a constant independent of $\bm u_t$.
Therefore, \eqref{eq:q_u_with_const} and \eqref{eq:q_u_goal} are equal except for the constant term.
Eq. \eqref{eq:q_u_with_const} was obtained by equivalent transformation of \eqref{eq:q_u}, and \eqref{eq:q_u_goal} was obtained by equivalent transformation of \eqref{q_star_u}.
Thus, we have shown that \eqref{eq:q_u} is an update formula for computing \eqref{q_star_u}.
\end{IEEEproof}

To describe Lemmas \ref{lemma_q_z_T} and \ref{lemma_q_theta_tau_pi_beta}, we introduce the following definitions to simplify the notation:
\begin{align}
    & q(s \preceq \mathsf{s}(\bm u_t)) \coloneqq \mathbb{E}_{q(\bm u_t)}[I \{ s \preceq \mathsf{s}(\bm u_t) \} ], \label{q_s_t}\\
    & \bm Q_s \coloneqq \mathrm{diag} \{ q(s \preceq \mathsf{s}(\bm u_1)), q(s \preceq \mathsf{s}(\bm u_2)), \ldots, q(s \preceq \mathsf{s}(\bm u_{n})) \}. \label{Q_s}
\end{align}

\begin{Lemma}\label{lemma_q_z_T}
The posterior $q(\bm z, T)$ can be factorized as $q(T) \prod_{s \in \mathcal{L}_T} q(\bm z_s)$.
The update formula of $q(\bm z_s)$ is given as
\begin{align}
    q(\bm z_s) = \prod_{k=1}^K (\pi'_{s,k})^{z_{s,k}},
\end{align}
where
\begin{align}
    \pi'_{s,k} &\coloneqq \frac{\rho_{s,k}}{\sum_{k=1}^K \rho_{s,k}} \label{lemma_pi}
\end{align}
and
\begin{align}
    &\ln \rho_{s,k} \coloneqq \mathbb{E}_{q(\bm \pi)}[ \ln \pi_k] \nonumber \\
    &\quad + \sum_{t=1}^{n} q(s \preceq \mathsf{s}(\bm u_t)) \mathbb{E}_{q(\bm \theta, \bm \tau)} [\ln \mathcal{N}(x_t | \tilde{\bm x}_t^\top \bm \theta_k, \tau_k^{-1})]. \label{eq:rho_closed_form}
\end{align}

Also, the update formula of $q(T)$ is given as
\begin{align}
    q(T) = \left( \prod_{s \in \mathcal{I}_T} g'_s \right) \left( \prod_{s \in \mathcal{L}_T} (1-g'_s) \right),
\end{align}
where
\begin{align}
    g'_s &\coloneqq \! \begin{cases}
        \frac{g_s \prod_{s_\mathrm{ch} \in \mathrm{Ch}(s)} \phi_{s_\mathrm{ch}}}{\phi_s}, & s \in \mathcal{I}_\mathrm{max}, \\
        0, & s \in \mathcal{L}_\mathrm{max},
    \end{cases} \label{lemma_g} \\
    \phi_s &\coloneqq \! \begin{cases}
        (1-g_s) \sum_{k=1}^K \rho_{s,k} + g_s \prod_{s_\mathrm{ch} \in \mathrm{Ch}(s)} \phi_{s_\mathrm{ch}}, & \! \! s \in \mathcal{I}_\mathrm{max}, \\
        \sum_{k=1}^K \rho_{s,k}, & \! \! s \in \mathcal{L}_\mathrm{max}.
    \end{cases}
\end{align}
\end{Lemma}

\begin{IEEEproof}
Calculating the right-hand side of \eqref{q_star_z_T} and substituting \eqref{q_s_t}, we obtain
\begin{align}
    &\ln q(\bm z, T) \nonumber \\
    &= \ln p(T) + \mathbb{E}_{q(\bm \pi)}[\ln p(\bm z | \bm \pi, T)] \nonumber \\
    &\quad + \mathbb{E}_{q(\bm \theta, \bm \tau)}[\ln p(\bm x | \bm u, \bm z, \bm \theta, \bm \tau, T)] + \mathrm{const.} \\
    &= \sum_{s \in \mathcal{I}_T} \ln g_s + \sum_{s \in \mathcal{L}_T} \ln (1-g_s) \nonumber \\
    &\quad + \sum_{s \in \mathcal{L}_T} \sum_{t=1}^n q(s \preceq \mathsf{s}(\bm u_t)) \sum_{k=1}^K z_{s,k} \Bigl\{ \mathbb{E}_{q(\bm \pi)} [\ln \pi_k] \nonumber \\
    &\qquad \quad + \mathbb{E}_{q(\bm \theta, \bm \tau)} [\ln \mathcal{N}(x_t | \tilde{\bm x}_t^\top \bm \theta_k, \tau_k^{-1})] \Bigr\} + \mathrm{const.} \\
    &= \sum_{s \in \mathcal{I}_T} \ln g_s + \sum_{s \in \mathcal{L}_T} \ln (1-g_s) \nonumber \\
    &\qquad + \sum_{s \in \mathcal{L}_T} \ln \sum_{k=1}^K \rho_{s,k} + \sum_{s \in \mathcal{L}_T} \sum_{k=1}^K z_{s,k} \ln \pi'_{s,k} + \mathrm{const.}, \label{q_z_T_form}
\end{align}
where $\bm \rho_{s,k}$ and $\pi'_{s,k}$ are defined as in \eqref{eq:rho_closed_form} and \eqref{lemma_pi}, respectively.

From \eqref{q_z_T_form}, we see that $q(\bm z, T)$ can be factorized as
\begin{align}
    q(\bm z, T) = q(T) \prod_{s \in \mathcal{L}_T} q(\bm z_s),
\end{align}
and $q(\bm z_s)$ is given as
\begin{align}
    q(\bm z_s) = \prod_{k=1}^K (\pi'_{s,k})^{z_{s,k}}.
\end{align}

Furthermore, $q(T)$ corresponds to the posterior distribution when considering $\left(\prod_{s \in \mathcal{I}_T} g_s \right) \left(\prod_{s \in \mathcal{L}_T}(1-g_s) \right)$ as the prior distribution and $\prod_{s \in \mathcal{L}_T} \sum_{k=1}^K \rho_{s,k}$ as the likelihood function. From \cite[Theorem 7]{full_rooted_trees}, such posterior distribution can be represented as
\begin{align}
    q(T) = \left( \prod_{s \in \mathcal{I}_T} g'_s \right) \left( \prod_{s \in \mathcal{L}_T} (1-g'_s) \right),
\end{align}
where $g'_s$ is defined as in \eqref{lemma_g}.
\end{IEEEproof}

To describe Lemma \ref{lemma_q_theta_tau_pi_beta}, we introduce the following definition to simplify the notation:
\begin{align}
    q(s \in \mathcal{L}_T) \coloneqq \mathbb{E}_{q(T)}[I \{ s \in \mathcal{L}_T \}]. \label{q_s_L}
\end{align}
Also, it should be noted that $\mathbb{E}_{q(\bm z_s)}[z_{s,k}] = \pi'_{s,k}$ holds because the $q(\bm z_s)$ is the categorical distribution with parameter $\bm \pi'_s$.

\begin{Lemma}\label{lemma_q_theta_tau_pi_beta}
The posterior $q(\bm \theta, \bm \tau, \bm \pi, \bm \beta)$ can be factorized as $q(\bm \pi) \prod_{k=1}^K q(\bm \theta_k, \tau_k) \prod_{s \in \mathcal{I}_\mathrm{max}} q(\bm \beta_s)$. The update formula of $q(\bm \pi)$ is given as
\begin{align}
    q(\bm \pi) = \mathrm{Dir}(\bm \pi | \bm \alpha'), \label{eq:q_pi}
\end{align}
where each element of $\bm \alpha'$ is given as
\begin{align}
    \alpha'_k \coloneqq \alpha_k + \sum_{s \in \mathcal{S}_\mathrm{max}} q(s \in \mathcal{L}_T) \pi'_{s,k}.
\end{align}

Also, $q(\bm \theta_k, \tau_k)$ is given as
\begin{align}
    q(\bm \theta_k, \tau_k) = \mathcal{N}(\bm \theta_k | \bm \mu'_k, (\tau_k \bm \Lambda'_k)^{-1}) \mathrm{Gam}(\tau_k | a'_k, b'_k), \label{eq:q_theta_tau}
\end{align}
where
\begin{align}
    \bm \Lambda'_k &\coloneqq \bm \Lambda + \sum_{s \in \mathcal{S}_\mathrm{max}} q(s \in \mathcal{L}_T) \pi'_{s,k} \bm X^\top \bm Q_s \bm X, \label{eq:Lambda_prime} \\
    \bm \mu'_k &\coloneqq \left( \bm \Lambda'_k \right)^{-1} \left( \bm \Lambda \bm \mu + \sum_{s \in \mathcal{S}_\mathrm{max}} q(s \in \mathcal{L}_T) \pi'_{s,k} \bm X^\top \bm Q_s \bm x \right), \label{eq:mu_prime} \\
    a'_k &\coloneqq a + \frac{1}{2}\sum_{s \in \mathcal{S}_\mathrm{max}} q(s \in \mathcal{L}_T) \pi'_{s,k} \mathrm{Tr} \{ \bm Q_s \}, \label{eq:a_prime} \\
    b'_k &\coloneqq b + \frac{1}{2} \Big( \bm \mu^\top \bm \Lambda \bm \mu + \sum_{s \in \mathcal{S}_\mathrm{max}} q(s \in \mathcal{L}_T) \pi'_{s,k} \bm x^\top \bm Q_s \bm x \nonumber \\
    &\qquad \qquad \qquad \qquad \qquad \qquad - (\bm \mu'_k)^\top \bm \Lambda'_k \bm \mu'_k \Big). \label{eq:b_prime}
\end{align}

Lastly, $q(\bm \beta_s)$ is given as
\begin{align}
    q(\bm \beta_s) = \mathcal{N}(\bm \beta_s | \bm \eta_s^\prime, (\bm L_s^\prime)^{-1}), \label{eq:q_beta}
\end{align}
where
\begin{align}
    \bm L_s^\prime &\coloneqq \bm L_s + 2 \sum_{t=1}^{n} q(s \preceq \mathsf{s}(\bm u_t)) \lambda (\xi_{s,t}) \tilde{\bm t} \tilde{\bm t}^\top, \\
    \bm \eta_s^\prime &\coloneqq \left( \bm L_s^\prime \right)^{-1} \left( \bm L_s \bm \eta_s + \sum_{t=1}^{n} q(s \preceq \mathsf{s}(\bm u_t)) \left( \varpi'_{t,s,s_1} - \frac{1}{2} \right) \tilde{\bm t} \right),
\end{align}
and the update formula of $\xi_{s,t}$ is
\begin{align}
    \xi_{s,t} =  (\tilde{\bm t}^\top \left( (\bm L_s^\prime)^{-1} + \bm \eta_s^\prime (\bm \eta_s^\prime)^\top \right) \tilde{\bm t})^{1/2}. \label{eq:xi_s_t}
\end{align}
\end{Lemma}

\begin{IEEEproof}
Calculating the right-hand side of \eqref{q_star_theta_tau_pi_beta}, we obtain
\begin{align}
    &\ln q (\bm \theta, \bm \tau , \bm \pi, \bm \beta) \nonumber \\
    &= \left( \ln p(\bm \pi) + \mathbb{E}_{q(\bm z)} [\ln p(\bm z | \bm \pi, T)] \right) \nonumber \\
    &\quad + \left( \ln p(\bm \theta, \bm \tau) + \mathbb{E}_{q(\bm u, \bm z, T)} [\ln p(\bm x | \bm u, \bm z, \bm \theta, \bm \tau, T)] \right) \nonumber \\
    &\quad + \left( \ln p(\bm \beta) + \mathbb{E}_{q(\bm u)} [\ln h(\bm u | \bm \beta, \bm \xi)] \right) + \mathrm{const.}
\end{align}
Therefore, $q(\bm \theta, \bm \tau, \bm \pi, \bm \beta)$ can be factorized as $q(\bm \pi) q(\bm \theta, \bm \tau) q(\bm \beta)$

For $q(\bm \pi)$, substituting \eqref{q_s_L}, we have
\begin{align}
    &\ln q(\bm \pi) \nonumber \\
    &= \ln \mathrm{Dir} (\bm \pi | \bm \alpha) + \mathbb{E}_{q(\bm z, T)} \left[ \sum_{s \in \mathcal{L}_T} \sum_{k=1}^K z_{s,k} \ln \pi_k \right] + \mathrm{const.} \\
    &= \sum_{k=1}^K (\alpha_k - 1) \ln \pi_k \nonumber \\
    &\quad +\mathbb{E}_{q(T)} \left[ \sum_{s \in \mathcal{S}_\mathrm{max}} I\{ s \in \mathcal{L}_T \} \sum_{k=1}^K \pi'_{s,k} \ln \pi_k \right] + \mathrm{const.} \\
    &= \sum_{k=1}^K (\alpha_k - 1) \ln \pi_k \nonumber \\
    &\quad + \sum_{s \in \mathcal{S}_\mathrm{max}} q(s \in \mathcal{L}_T) \sum_{k=1}^K \pi'_{s,k} \ln \pi_k+ \mathrm{const.} \label{end}\\
    &= \sum_{k=1}^K \left(\alpha_k + \sum_{s \in \mathcal{S}_\mathrm{max}} q(s \in \mathcal{L}_T) \pi'_{s,k} - 1 \right) \ln \pi_k + \mathrm{const.}
\end{align}
Therefore, we obtain \eqref{eq:q_pi}.

For $q(\bm \theta, \bm \tau)$, using \eqref{q_s_t} and \eqref{q_s_L}, we have
\begin{align}
    &\ln q(\bm \theta, \bm \tau) \nonumber \\
    &=\sum_{k=1}^K \ln \mathcal{N} (\bm \theta_k | \bm \mu, (\tau_k \bm \Lambda)^{-1}) \mathrm{Gam}(\tau_k | a, b) \nonumber \\
    &\quad + \mathbb{E}_{q(\bm u, \bm z, T)} \Biggl[ \sum_{t=1}^{n} \sum_{s \in \mathcal{L}_T} \sum_{k=1}^K z_{s,k} I\{s \preceq \mathsf{s}(\bm u_t)\} \nonumber \\
    &\qquad \qquad \qquad \qquad \times \ln \mathcal{N}(x_t | \tilde{\bm x}_t^\top \bm \theta_k, \tau_k^{-1}) \Biggr] + \mathrm{const.} \\
    &=\sum_{k=1}^K \ln \mathcal{N} (\bm \theta_k | \bm \mu, (\tau_k \bm \Lambda)^{-1}) \mathrm{Gam}(\tau_k | a, b) \nonumber \\
    &\quad + \mathbb{E}_{q(\bm u, \bm z, T)} \Biggl[ \sum_{t=1}^{n} \sum_{s \in \mathcal{S}_\mathrm{max}} \sum_{k=1}^K z_{s,k} I\{s \preceq \mathsf{s}(\bm u_t)\} I \{ s \in \mathcal{L}_T \} \nonumber \\
    &\qquad \qquad \qquad \qquad \times \ln \mathcal{N}(x_t | \tilde{\bm x}_t^\top \bm \theta_k, \tau_k^{-1}) \Biggr] + \mathrm{const.} \\
    &= \sum_{k=1}^K \ln \mathcal{N} (\bm \theta_k | \bm \mu, (\tau_k \bm \Lambda)^{-1}) \mathrm{Gam}(\tau_k | a, b) \nonumber \\
    &\quad + \sum_{t=1}^{n} \sum_{s \in \mathcal{S}_\mathrm{max}} \sum_{k=1}^K \pi'_{s,k} q(s \preceq \mathsf{s}(\bm u_t)) q(s \in \mathcal{L}_T) \nonumber \\
    & \qquad \qquad \qquad \qquad \times \ln \mathcal{N}(x_t | \tilde{\bm x}_t^\top \bm \theta_k, \tau_k^{-1})+ \mathrm{const.} \\
    &= \sum_{k=1}^K \Biggl\{ \ln \mathcal{N} (\bm \theta_k | \bm \mu, (\tau_k \bm \Lambda)^{-1}) \mathrm{Gam}(\tau_k | a, b) \nonumber \\
    &\quad + \sum_{s \in \mathcal{S}_\mathrm{max}} \frac{1}{2} q(s \in \mathcal{L}_T) \pi'_{s,k} \biggl( \mathrm{Tr}\{ \bm Q_s \} \ln \tau_k \nonumber \\
    & \qquad \quad - \tau_k (\bm x - \bm X\bm \theta_k)^\top \bm Q_s (\bm x - \bm X\bm \theta_k) \biggr) \Biggr\} + \mathrm{const.}
\end{align}
Therefore, $q(\bm \theta, \bm \tau)$ can be factorized as $\prod_{k=1}^K q(\bm \theta_k, \tau_k)$. For each $q(\bm \theta_k, \tau_k)$, by completing the square with respect to $\bm \theta_k$ and letting $\bm \Lambda'_k$, $\bm \mu'_k$, $a'_k$, and $b'_k$ be defined as \eqref{eq:Lambda_prime}, \eqref{eq:mu_prime}, \eqref{eq:a_prime}, and \eqref{eq:b_prime}, respectively, we get \eqref{eq:q_theta_tau}.

For $q(\bm \beta)$, substituting \eqref{eq_prior_beta} and \eqref{eq_def_h(u)} and rearranging the terms, we have
\begin{align}
    &\ln q(\bm \beta) 
    = \sum_{s \in \mathcal{I}_\mathrm{max}} \Big \{ -\frac{1}{2}(\bm \beta_s -\bm \eta_s)^\top \bm L_s (\bm \beta_s -\bm \eta_s) \nonumber \\
    &+\sum_{t=1}^n \mathbb{E}_{q(\bm u_t)} \Big [I \{ s \preceq \mathsf{s}(\bm u_t) \} \Big (\bm \beta_s^\top \tilde{\bm t} u_{t,d_s} -\frac{1}{2} (\bm \beta_s^\top \tilde{\bm t} + \xi_{s,t}) \nonumber \\
    & \qquad \qquad \qquad - \lambda(\xi_{s,t}) ((\bm \beta_s^\top \tilde{\bm t})^2 - \xi_{s,t}^2 ) \Big) \Big] \Big \}+ \mathrm{const.} 
\end{align}
By completing the square with respect to $\bm \beta_s$, we obtain \eqref{eq:q_beta}. Also, \eqref{eq:xi_s_t} follows from the same argument as in (10.161)--(10.163) in \cite{bishop}.
\end{IEEEproof}

\begin{Lemma}\label{lemma_expectations}
The expectations in Lemmas \ref{lemma_q_u}, \ref{lemma_q_z_T}, and \ref{lemma_q_theta_tau_pi_beta} can be calculated as follows.
\begin{align}
    &\mathbb{E}_{q(\bm u_t)}[I \{ s \preceq \mathsf{s}(\bm u_t) \} ] = \prod_{s', s'_\mathrm{ch} \preceq s} \varpi'_{t, s', s'_\mathrm{ch}}, \label{calc_q_s_t}\\
    &\mathbb{E}_{q(T)}[I \{ s \in \mathcal{L}_T \}] = (1-g'_s) \prod_{s' \prec s} g'_{s'}, \label{calc_q_s_L} \\
    &\mathbb{E}_{q(\bm \beta_s)}[ \ln h( u | \bm \beta_s, \xi_{s,t})] \nonumber \\
        &\quad = \ln \sigma (\xi_{s,t}) + (\bm \eta_s^\prime)^\top \tilde{\bm t} u -\frac{1}{2} \left( (\bm \eta_s^\prime)^\top \tilde{\bm t} + \xi_{s,t} \right) \nonumber \\
        &\qquad - \lambda(\xi_{s,t}) \Bigl( \tilde{\bm t}^\top (\bm \eta_s^\prime (\bm \eta_s^\prime)^\top + (\bm L_s^\prime)^{-1}) \tilde{\bm t} - (\xi_{s,t})^2 \Bigr), \label{calc_E_h_u}\\
    &\mathbb{E}_{q(\bm z, T, \bm \theta, \bm \tau)} \left[I \{ s_u \in \mathcal{L}_T \} \sum_{k=1}^K z_{s_u, k} \ln \mathcal{N}(x_t | \tilde{\bm x}_t^\top \bm \theta_k, \tau_k^{-1})\right] \nonumber \\
        &\quad = \frac{1}{2} (1-g'_{s_u}) \left( \prod_{s \prec s_u} g'_{s} \right) \sum_{k=1}^K \pi'_{s_u,k} \nonumber \\
        &\qquad \times \biggl\{ -\ln 2\pi + \psi(a'_k) - \ln b'_k \nonumber \\
        &\qquad \qquad - \frac{a'_k}{b'_k}(x_t - \tilde{\bm x}_t^\top \bm \mu'_k)^2 - \tilde{\bm x}_t^\top (\bm \Lambda'_k)^{-1} \tilde{\bm x}_t \biggr\}, \label{calc_E_ln_p_x_for_q_u}\\
    & \mathbb{E}_{q(\bm \pi)}[ \ln \pi_k] = \psi (\alpha'_k) - \psi \left( \sum_{k=1}^K \alpha'_k \right), \label{calc_E_ln_pi} \\
    &\sum_{t=1}^{n} q(s \preceq \mathsf{s}(\bm u_t)) \mathbb{E}_{q(\bm \theta, \bm \tau)} [\ln \mathcal{N}(x_t | \tilde{\bm x}_t^\top \bm \theta_k, \tau_k^{-1})] \nonumber \\
    & \quad =  \frac{1}{2} \mathrm{Tr} \{ \bm Q_s \} \left( -\ln 2\pi + \psi (a'_k) - \ln b'_k \right) \nonumber \\
    &\qquad - \frac{a'_k}{2 b'_k} (\bm x - \bm X \bm \mu'_k)^\top \bm Q_s (\bm x - \bm X \bm \mu'_k) \nonumber \\
    & \qquad -\frac{1}{2} \mathrm{Tr} \{ \bm X^\top \bm Q_s \bm X (\bm \Lambda'_k)^{-1} \}. \label{calc_E_ln_p_x_for_q_z}
\end{align}
\end{Lemma}

\begin{IEEEproof}
First, we calculate $\mathbb{E}_{q(\bm u_t)} [I \{ s \preceq \mathsf{s}(\bm u_t) \}]$, which is also denoted as $q(s \preceq \mathsf{s}(\bm u_t))$ in Appendix or $q_{s,t}$ in the main article. Qualitatively, $\varpi'_{t,s',s'_\mathrm{ch}}$ in $q(\bm u_t)$ denotes the probability that the path going through $s'$ takes the branch to $s'_\mathrm{ch}$. Since $\mathbb{E}_{q(\bm u_t)} [I \{ s \preceq \mathsf{s}(\bm u_t) \}]$ denotes the probability that $\mathsf{s}(\bm u_t)$ is in the descendant nodes of $s$, it can be calculated as the product of $\varpi'_{t, s', s'_\mathrm{ch}}$ for all the pair of $(s', s'_\mathrm{ch})$ on the path from $s_\lambda$ to $s$. Thus, \eqref{calc_q_s_t} holds.

Next, the expectation $\mathbb{E}_{q(T)} [I \{ s \in \mathcal{L}_T \}]$, which is also denoted as $q(s \in \mathcal{L}_T)$, is given as \eqref{calc_q_s_L} from \cite[Theorem 2]{full_rooted_trees}.

For $\mathbb{E}_{q(\bm \beta_s)}[\ln h(u | \bm \beta_s, \xi_{s,t})]$, since $q(\bm \beta_s)$ is the normal distribution $\mathcal{N}(\bm \beta_s | \bm \eta'_s, (\bm L'_s)^{-1})$, \eqref{calc_E_h_u} straightforwardly holds by using the properties of the mean vector $\bm \eta'_s = \mathbb{E}_{q(\bm \beta_s)}[\bm \beta_s]$ and the covariance matrix $(\bm L'_s)^{-1} = \mathbb{E}_{q(\bm \beta_s)}[ \bm \beta_s \bm \beta_s^\top] - \bm \eta'_s (\bm \eta'_s)^\top$.

For \eqref{calc_E_ln_p_x_for_q_u}, substituting \eqref{calc_q_s_L} and the expectation of the categorical distribution $\mathbb{E}_{q(\bm z_s)}[z_{s,k}] = \pi'_{s,k}$, we have
\begin{align}
    &\mathbb{E}_{q(\bm z, T, \bm \theta, \bm \tau)} \left[I \{ s_u \in \mathcal{L}_T \} \sum_{k=1}^K z_{s_u, k} \ln \mathcal{N}(x_t | \tilde{\bm x}_t^\top \bm \theta_k, \tau_k^{-1})\right] \nonumber \\
    &= (1-g'_{s_u}) \left( \prod_{s \prec s_u} g'_s \right) \sum_{k=1}^K \pi'_{s_u, k} \nonumber \\
    &\qquad \qquad \qquad \quad \times \mathbb{E}_{q(\bm \theta_k, \tau_k)} \left[ \ln \mathcal{N}(x_t | \tilde{\bm x}_t^\top \bm \theta_k, \tau_k^{-1})\right].
\end{align}
Further, since $q(\tau_k)$ is the gamma distribution $\mathrm{Gam}(\tau_k | a'_k, b'_k)$, $\mathbb{E}_{q(\tau_k)}[\tau_k]=\frac{a'_k}{b'_k}$ and $\mathbb{E}_{q(\tau_k)}[\ln \tau_k] = \psi(a'_k) - \ln b'_k$ hold. Moreover, since $q(\bm \theta_k | \tau_k)$ is the normal distribution with mean $\bm \mu'_k$ and covariance matrix $(\tau_k \bm \Lambda'_k)^{-1}$, we have $\bm \mu'_k = \mathbb{E}_{q(\bm \theta_k | \tau_k)}[\bm \theta_k]$ and $(\tau_k \bm \Lambda'_k)^{-1} = \mathbb{E}_{q(\bm \theta_k | \tau_k)}[\bm \theta_k \bm \theta_k^\top] - \bm \mu'_k (\bm \mu'_k)^\top$. Therefore, we obtain
\begin{align}
    &\mathbb{E}_{q(\bm \theta_k, \tau_k)}[\ln \mathcal{N}(x_t | \tilde{\bm x}_t^\top \bm \theta_k, \tau_k^{-1})] \nonumber \\
    &= \frac{1}{2} \biggl\{ -\ln 2\pi + \psi(a'_k) - \ln b'_k \nonumber \\
    &\qquad \quad  - \frac{a'_k}{b'_k}(x_t - \tilde{\bm x}_t^\top \bm \mu'_k)^2 - \tilde{\bm x}_t^\top (\bm \Lambda'_k)^{-1} \tilde{\bm x}_t \biggr\}. \label{E_ln_p_x}
\end{align}
Consequently, \eqref{calc_E_ln_p_x_for_q_u} holds.

Equation \eqref{calc_E_ln_pi} is a well known property of the Dirichlet distribution. Lastly, using \eqref{Q_s} and \eqref{E_ln_p_x}, and representing the sum with respect to $t$ as matrix products, we obtain \eqref{calc_E_ln_p_x_for_q_z}.
\end{IEEEproof}

\end{document}